\documentclass[runningheads]{llncs}
\usepackage[misc]{ifsym}

\usepackage{graphicx}
\usepackage{amsmath}
\usepackage{pgf-pie}
\usepackage[utf8]{inputenc}
\usepackage{verbatim}
\usepackage{cite}
\usepackage[english]{babel}
\usepackage[noend]{algpseudocode}
\usepackage{array}
\usepackage{graphicx}
\usepackage[small]{caption}
\usepackage{enumitem}
\usepackage{graphics}
\usepackage{color}
\usepackage{adjustbox}
\usepackage{float}
\usepackage{url}
\usepackage{hyperref}
\usepackage{amssymb}

\usepackage{algorithm}

\begin{document}
\title{Obj2Sub: Unsupervised Conversion of Objective to Subjective Questions\thanks{This research work is supported by Extramarks Education India Pvt. Ltd., SERB, FICCI (PM fellowship), Infosys Centre for AI and TiH Anubhuti (IIITD).}}
\author{Aarish Chhabra \Letter \and Nandini Bansal \and Venktesh V \and Mukesh Mohania \and Deep Dwivedi}
\tocauthor{Aarish ~ Chhabra, Nandini ~ Bansal, Venktesh ~ V, Mukesh ~ Mohania, Deep Dwivedi}
\toctitle{K-12BERT: BERT for K-12 education}
\authorrunning{Aarish et. al.}
%
\institute{Indraprastha Institute of Information Technology, Delhi \email{\{aarish17212,nandini18056,venkteshv,mukesh, deepd\}@iiitd.ac.in}}
%
%
%
\maketitle              
\begin{abstract}
Exams are conducted to test the learner's understanding of the subject. To prevent the learners from guessing or exchanging solutions, the mode of tests administered must have sufficient subjective questions that can gauge whether the learner has understood the concept by mandating a detailed answer. Hence, in this paper, we propose a novel hybrid unsupervised approach leveraging rule based methods and pre-trained dense retrievers for the novel task of automatically converting the objective questions to subjective questions. We observe that our approach outperforms the existing data-driven approaches by \textbf{36.45\%} as measured by Recall@k and Precision@k.
\end{abstract}
\section{Introduction and Related Work}

In online platforms, assessments are critical to gauge if the learner has understood the concept. However, assessments with only objective questions may prompt the user to just guess the answer using the options. For more rigorous testing of understanding,we propose the novel task of converting objective questions to subjective questions to mandate a detailed answer. Let Q be an objective question (OQ) and A be the answer to that objective question. Our task is to convert Q to a short subjective question (SQ) S. For example, let's say \textit{Q: The wastes that can choke the drains include}, \textit{A: used tea leaves, cotton}, then the possible subjective question could be \textit{S: What kind of wastes can choke the drains?} The task of converting objective questions to short subjective questions in the absence of labeled pairs (OQ-SQ pairs) hasn't been specifically explored previously to the best of our knowledge.

Many state-of-the-art Question Generation (QG) systems have been proposed in recent years  \cite{Luetl,Kim_Lee_Shin_Jung_2019,KHODEIR201410}. These systems usually use deep learning-based approaches such as Seq2Seq models \cite{sutskever2014sequence} or more recently, transformers \cite{vaswani2017attention}. However, the mentioned approaches have complex model architectures and require significant amounts of labeled data for training.

\begin{figure*}
\includegraphics[scale=0.4]{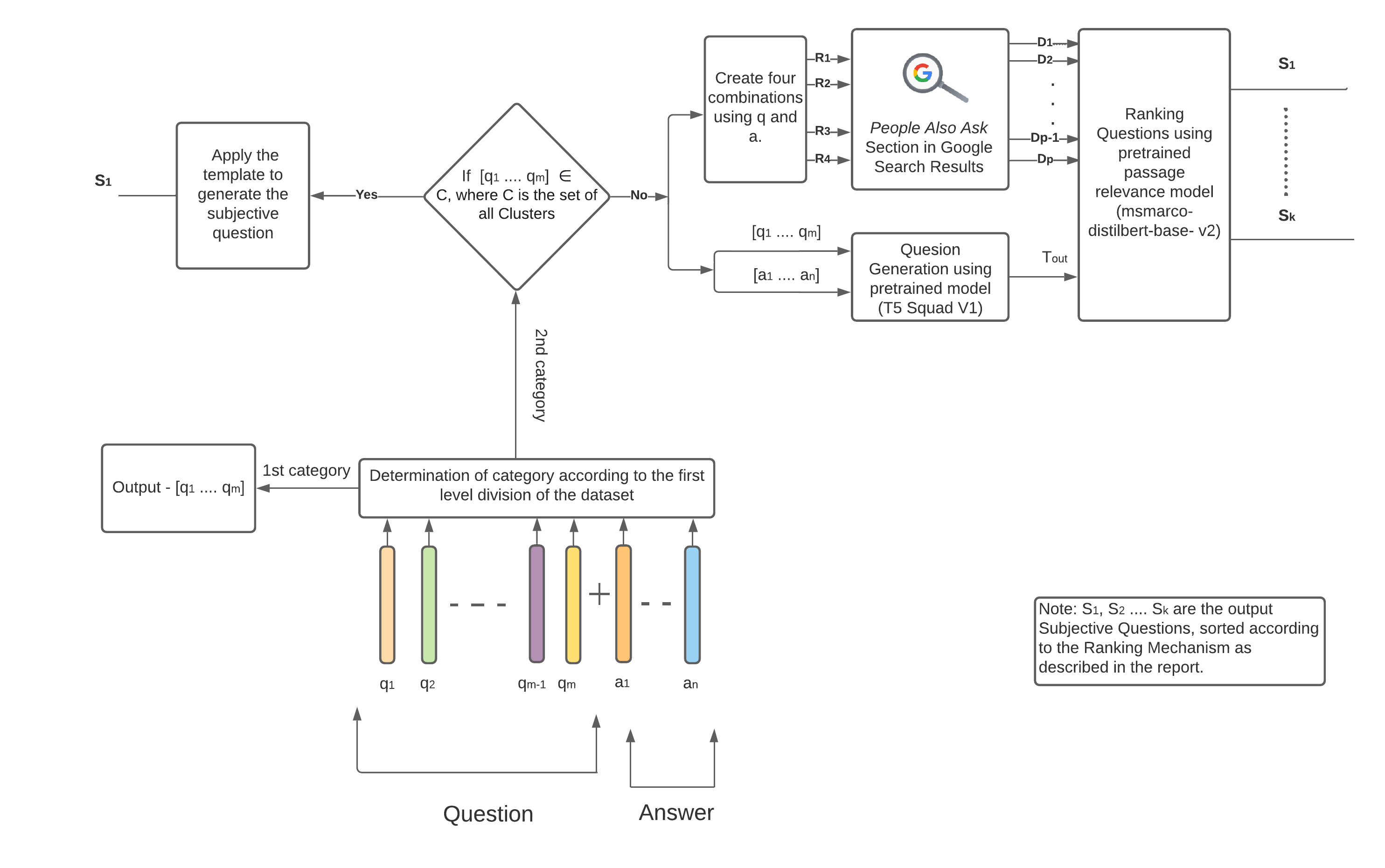}
\caption{Architecture for conversion of objective question to a subjective question} \label{architecture}
\end{figure*}
We open-source our code and datasets at \url{https://github.com/ADS-AI/Obj2Sub-AIED2022}

\section{Methodology}
In this section, we discuss the proposed unsupervised method for generating subjective question(s) from a given objective question. Let Q be a given objective question, A be an answer to this objective question, q\textsubscript{1}q\textsubscript{2}... q\textsubscript{m} and a\textsubscript{1}a\textsubscript{2} ... a\textsubscript{n} be the token sequence of the objective question and answer respectively. Let S\textsubscript{1} .... S\textsubscript{k} be the subjective questions generated by the proposed Obj2Sub method for a given \textless Q, A\textgreater \ pair. 
We propose a \textit{novel}, \textit{unsupervised} and a \textit{hybrid} approach to automate the process of converting an objective question into a subjective question. Fig. \ref{architecture} gives the complete picture of the proposed methodology. Upon receiving the inputs (Q and A), our system automatically classifies Q into one of the 3 broad categories of objective questions based on what class it represents:
\begin{itemize}
    \item \textbf{Multi-option Dependent} ($\sim$ 7\%) The class of questions that are dependent on the objective question's options and don't focus on a single learning concept. This category has not been dealt with in our proposed method because these are negligible in number. These can be easily filtered using the presence of phrases such as \textit{of the following}, \textit{choose the statement}, etc.
    \item \textbf{WhWord} ($\sim$ 61\%) The class of questions that can be answered without looking into the option and can be directly used as subjective questions. These can be easily identified using the presence of wh-words as a first token ($q_1$) in Q.

    \item \textbf{Declarative Sentence} ($\sim$ 32\%) This category contains the set of objective questions in which Q + A (’+’ depicts concatenation) forms a declarative sentence. The questions not filtered out using either of the first 2 steps are mostly observed to be falling under this category. Eg: Question = The chemical symbol for silver is, Answer = Ag
\end{itemize}

\noindent In this paper, we focus on the conversion of Objective Questions belonging to Declarative Sentence category since this is a more common category of objective questions. For the conversion, we follow a hybrid approach which consists of 3 major components:

\begin{itemize}
    \item \textbf{Clustering and Rule-Based Templates} We observe that some specific tokens are the same for multiple objective questions. These tokens are generally either the \textit{last token ($q_m$), or the last two tokens ($q_m$ , $q_{m-1}$), or the first token ($q_1$)}. For example, the presence of \textbf{by} as the last token can be seen in the objective questions: \textit{Law of constant proportions is given \textbf{by}, Polio is caused \textbf{by}}. Thus, we define various clusters of objective questions based on the presence of these specific tokens. However, all these clusters don't represent a broad class of objective questions, leading us to perform cluster pruning based on the frequency (\textit{\textgreater 500}) in our dataset. Further, we define a single rule-based template using various syntactic features such as part of speech tags, determining auxiliary verbs, named entity
recognition of tokens, changing verb forms (lemmatization), subject-auxiliary-inversion as defined in \cite{heilman-smith-2010-good} to steer the conversion into a short subjective question. Questions not covered with clustering are dealt using the other 2 components.
    \item \textbf{Leveraging Open Source Knowledge Base} We utilize open-source knowledge base such as the People Also Ask (PAA) section of Google for the conversion task. We form multiple search queries using the input tokens (Q, A) and extract top 4 PAA questions using APIs of a Python-based library. The problem of inconsistency in search results is tackled smartly by using different permutations for forming search queries and devising a filtering mechanism to discard irrelevant questions. Eg: Q = desert plants have scale/spine-like leaves to, A = reduce the loss of water by transpiration, a sample S = How are the desert plants adapted to reduce the loss of water by transpiration?

    \item \textbf{Pre-Trained T5 based model} To eliminate a rare problem of concept drift in the questions generated using PAA section, we utilize a transformer-based architecture (T5 Squad V1 \footnote{\url{https://huggingface.co/ramsrigouthamg/t5_squad_v1}}) to augment the generated questions further by conditioning the generation of new subjective questions on the given
    
\begin{table*}[hbt !]
\centering
\caption{R@k and P@k for k=1,2,3 on ObjQA and MCQ Datasets}\label{Metrics}
\begin{tabular}{|p{0.09\linewidth}|p{0.30\linewidth}|p{0.08\linewidth}|p{0.1\linewidth}|p{0.1\linewidth}|p{0.09\linewidth}|p{0.09\linewidth}|p{0.09\linewidth}|}

\hline
Dataset & Method & R@1 & R@2 & R@3 & P@1 & P@2 & P@3 \\
\hline

 & \textbf{Obj2Sub}(our method) & \bf0.203 & \bf0.318 & \bf0.408 & \bf0.610 & \bf0.477 & \bf0.408 \vspace{0.25cm}
\\

ObjQA & Rule-Based Approach \cite{heilman-smith-2010-good} & 0.110 & 0.189 & 0.222 & 0.332 & 0.283 & 0.222
\\

 & T5-Transformer \cite{raffel2020exploring} & 0.183 & 0.246 & 0.299 & 0.550 & 0.370 & 0.299\\
\hline
 & \textbf{Obj2Sub}(our method) & \bf0.255 & \bf0.329 & \bf0.393 & \bf0.767 & \bf0.493 & \bf0.393 \vspace{0.25cm}
\\

MCQ & Rule-Based Approach \cite{heilman-smith-2010-good} & 0.156 & 0.276 & 0.317 & 0.47 & 0.415 & 0.317
\\

 & T5-Transformer \cite{raffel2020exploring} & 0.195 & 0.292 & 0.378 & 0.586 & 0.439 & 0.378\\

\hline
\end{tabular}
\end{table*}

objective question and the context (the answer to the Objective Question) around which the question must be framed.
\end{itemize}

\noindent Now, we have a set S of subjective questions generated using the last 2 components discussed above. We further rank the questions using a pre-trained ranking model (\textit{msmarco-distilroberta-base-v2}\footnote{\url{https://huggingface.co/sentence-transformers/msmarco-distilroberta-base-v2}}) and fetch top-k questions. For this paper, we mostly stick to k=3.

\section{Experiments and Results}
In this paper, we compare our results with two different kinds of dataset for a
holistic evaluation of the proposed system.

\noindent \textbf{ObjQA Dataset}: This is a proprietary dataset from an e-learning platform
which consists of approx 2,70,000 non-visual K-12 based objective question samples spanning different subjects.

\noindent \textbf{MCQ Dataset}:  This is an open-source dataset which is used to further verify
the robustness of the system proposed in this paper. Again, questions from different subjects are picked to keep the experiments unbiased.

\vspace{1em}
\noindent Due to the novelty of the problem statement, we compare our results with 2 closely related question generation systems, \textbf{Transformer based T5 Squad V1} (taking the top-3 outputs) and \textbf{Rule Based Approach devised in \cite{heilman-smith-2010-good}}. Table 1 shows the various results and suggests that our method outperforms the existing methodologies by \textbf{36.45\%} as measured using Recall@3 and Precision@3 due to the hybrid nature of the approach.







%
%
%
\bibliographystyle{splncs04}
\bibliography{AIED}

%




\end{document}